\documentclass[conference, 10pt, final]{IEEEtran}
    
\IEEEoverridecommandlockouts

\usepackage{cite}
\usepackage{algorithmic}
\usepackage{graphicx}
\graphicspath{ {./Figures/} }
\usepackage{textcomp}
\usepackage{xcolor}

\def\BibTeX{{\rm B\kern-.05em{\sc i\kern-.025em b}\kern-.08em
    T\kern-.1667em\lower.7ex\hbox{E}\kern-.125emX}}

\usepackage{amssymb, amsmath,amsfonts,amsthm,bm} % Math packages for equations
\usepackage{mathtools}
\usepackage{dsfont} % Used for bold 1 vector

\usepackage[utf8]{inputenc}
\usepackage[acronym]{glossaries}
\usepackage{subcaption}

\usepackage[ruled,vlined]{algorithm2e}
\SetKw{KwBy}{by}
\SetKwInOut{Input}{Input}
\SetKwInOut{Output}{Output}
\SetKwRepeat{Do}{do}{while}
\usepackage[T1]{fontenc}
\usepackage[utf8]{inputenc}

\usepackage{pgfplots}
\usepackage{tikz}
\usetikzlibrary{positioning}
\usetikzlibrary{shapes}
\usetikzlibrary{calc}

\usepackage{times}  % DO NOT CHANGE THIS
\usepackage{helvet} % DO NOT CHANGE THIS
\usepackage{courier}  % DO NOT CHANGE THIS
\usepackage[hyphens]{url}  % DO NOT CHANGE THIS
\usepackage{graphicx} % DO NOT CHANGE THIS
\usepackage{amsmath,amsfonts,amsthm} % Math packages for equations
\usepackage{mathtools}
\usetikzlibrary{positioning}
\usetikzlibrary{shapes}
\usetikzlibrary{calc}

\usepackage{pgf}
\usepackage{relsize}

\usepackage{pgfplots}
\pgfmathdeclarefunction{gauss}{2}{%
	\pgfmathparse{1/(#2*sqrt(2*pi))*exp(-((x-#1)^2)/(2*#2^2))}%
}

\tikzset{fontscale/.style = {font=\relsize{#1}}
}

\newacronym{mc}{MC}{Monte Carlo}
\newacronym{al}{AL}{Active Learning}
\newacronym{mmp}{MMP}{Mix-Marketing Plan}
\newacronym{rmse}{RMSE}{Root Mean Squared Error}
\newacronym{gan}{GAN}{Generative Adversarial Network}
\newacronym{sota}{SOTA}{state-of-the-art}
\newacronym{ml}{ML}{Machine Learning}
\newacronym{bm}{BM}{Bayesian Model}
\newacronym{mle}{MLE}{Maximum Likelihood Estimate}
\newacronym{map}{MAP}{Maximum a Posteriori}
\newacronym{mcmc}{MCMC}{Markov Chain Monte Carlo}
\newacronym{ohe}{OHE}{One-hot Encoding}
\newacronym{mse}{MSE}{Mean Squared Error}
\newacronym{nn}{NN}{Neural Network}
\newacronym{cnn}{CNN}{Convolutional Neural Network}
\newacronym{vi}{VI}{Variational Inference}
\newacronym{es}{ES}{Early Stopping}
\newacronym{pgm}{PGM}{Probabilistic Graphical Model}
\newacronym{vc}{VC}{Vapnik-Chervonenkis}
\newacronym{vae}{VAE}{Variational Autoencoder}

\begin{document}

\title{Rapid Risk Minimization with Bayesian Models Through Deep Learning Approximation
\thanks{This project was supported by the Innovation Fund Denmark (Grant ID 8053-00073B).}
}

\author{\IEEEauthorblockN{Mathias Löwe\IEEEauthorrefmark{1}, Per Lunnemann\IEEEauthorrefmark{2} and Sebastian Risi\IEEEauthorrefmark{1}}
	\IEEEauthorblockA{\IEEEauthorrefmark{1}	\textit{IT University of Copenhagen, Copenhagen, Denmark}}
	\IEEEauthorblockA{\IEEEauthorrefmark{2}\textit{Blackwood Seven, Copenhagen, Denmark}\\\textit{malw@itu.dk, lunnemann@gmail.com, sebr@itu.dk}}
}
\maketitle

\begin{abstract}
We introduce a novel combination of \glspl{bm} and \glspl{nn} for making predictions with a minimum expected risk. Our approach combines the best of both worlds, the data efficiency and interpretability of a \gls{bm} with the speed of a \gls{nn}. For a \gls{bm}, making predictions with the lowest expected loss requires integrating over the posterior distribution. When exact inference of the posterior predictive distribution is intractable, approximation methods are typically applied, e.g.\@ \gls{mc} simulation. For MC, the variance of the estimator decreases with the number of samples -- but at the expense of increased computational cost. 
Our approach removes the need for iterative \gls{mc} simulation on the CPU at prediction time. In brief, it works by fitting a NN to synthetic data generated using the \gls{bm}. In a single feed-forward pass, the \gls{nn} gives a set of point-wise approximations to the \gls{bm}'s posterior predictive distribution for a given observation. We achieve risk minimized predictions significantly faster than standard methods with a negligible loss on the test dataset. We combine this approach with \gls{al} to minimize the amount of data required for fitting the \gls{nn}. This is done by iteratively labeling more data in regions with high predictive uncertainty of the \gls{nn}.
\end{abstract}

\begin{IEEEkeywords}
Bayesian Models, Neural Networks, Active Learning, Bayes' risk
\end{IEEEkeywords}

\glsresetall
\section{Introduction}
The toolbox of machine learning is ever-growing and contains a wide spectrum of tools. Each new tool has its advantages and drawbacks. \glspl{nn}, for instance, are \emph{model free}, universal approximators capable of learning complex characteristics and generally allow for making fast predictions once fitted. Their drawbacks are, among others, the need for a vast amount of data, difficulty to debug, and lack of interpretability. At the other end of the spectrum, we have \emph{model based} methods like \glspl{pgm}. \Glspl{pgm} are handcrafted models describing the process of \emph{how} the observed data arose. To learn the parameters of these models, \emph{Bayesian inference} can be applied through the use of Bayes' theorem. Throughout this paper, we will refer to such constructs as \glspl{bm}. \Glspl{bm} are generative, meaning it is possible to generate synthetic data through sampling \cite{deisenroth2020mathematics}. Unlike \glspl{nn}, \glspl{bm} are data-efficient and have integrated uncertainty handling. This is at the cost of computational complexity, this being one of their main drawbacks. There are two computational issues with \glspl{bm} 1) obtaining the posterior distribution over the free parameters, and 2) making predictions using the posterior predictive distribution. With the rise of sophisticated techniques for posterior inference like \gls{vi} and \gls{mcmc}, \glspl{bm} have gained an increasing level of popularity over the past decades. In contrast to the former computational issue, the latter has gained little attention in academic research. 

The core idea of this paper is taking the best of \glspl{nn} and \glspl{bm} to combine them for solving this issue through approximation. The proposed method allows one to obtain a fast approximation of the posterior predictive. \emph{In brief, it works by fitting a \gls{nn} to synthetic data in which each prediction gives a set of point-wise approximations of the \gls{bm}'s mean posterior predictive distribution}. The method is especially useful in situations where a large number of predictions are needed, e.g.\@ when optimizing over the posterior predictive. It utilizes each tool's respective strengths to counter the weaknesses of the other, in particular the fast predictions of a \gls{nn} with the data-efficiency of the \gls{bm}. 
We present an algorithm on how to perform the combination of the two methods, how the approximation is conducted, and how the \gls{nn} can be used as a surrogate function to enable fast, accurate, predictions. As our focus is not on obtaining the posterior distribution, we will assume a well-performing \gls{bm} with parameters $\phi$ on some dataset $\mathcal{D}^\text{\gls{bm}}$ and with a well-approximated posterior distribution, $p\left(\phi \mid \mathcal{D}^\text{\gls{bm}} \right)$.

In the following, we will first introduce some concepts from basic decision theory to define how one makes the best possible prediction given some input, hence minimizing the expected risk. This will serve as a key to understanding why making risk minimized predictions with \glspl{bm} can be computationally heavy. Then, we present existing work in the field of approximating \glspl{nn} to \glspl{bm}. This will be followed by an introduction of the general method proposed in this paper and how it differs from related work. We do a formal analysis of the computational complexity involved in making predictions using our proposed method versus standard approaches, and present empirical results.

\subsection{Minimizing Risk} 
Assume we can obtain pairs of data samples, $\left(x, y\right)$, from some unknown joint probability distribution, $p\left(X, Y\right)$. Now, we are given an observation $x$ and intend to predict the correct corresponding, $y$. We will denote this prediction, $\widetilde{y}$.
Basic decision theory establish how one makes the best prediction in such a case by minimizing the expected loss $\mathbb{E}\left[\mathcal{L}\left(\widetilde{y}, y \right) \right]$ which is risk minimization \cite{deisenroth2020mathematics, bishop2006pattern}. Throughout this paper, we are interested in regression problems and will assume a Euclidean loss function, $\mathcal{L}\left(\widetilde{y}, y\right) = \left(\widetilde{y} - y \right)^2$. It can be shown that the prediction, $\widetilde{y}$, which minimizes this expected loss is $\mathbb{E}\left[Y \mid X \right]$ \cite{bishop2006pattern}.
Hence, the optimal prediction is the conditional expectation of the underlying data distribution. For other choices on the form of $\mathcal{L}$, the optimal prediction would be different \cite{mcelreath2020statistical}. An oracle predicting $\widetilde{y} = \mathbb{E}\left[Y \mid X \right]$, would still incur some error due to the stochasticity of the data generation process, this error is sometimes called Bayes risk or Bayes error \cite{goodfellow2016deep, robert2007bayesian}. Without access to an oracle, the ground true data distribution for most interesting problems is unknown, and we cannot compute the conditional expectation directly. Instead, one can create a model of it. If we assume a \gls{bm} with a posterior distribution $p\left(\phi \mid \mathcal{D}^\text{BM} \right) \propto p\left(\mathcal{D}^\text{BM} \mid \phi \right) p\left(\phi\right)$, we can use $\mathbb{E}\left[\widetilde{Y} \mid X \right]$ as a surrogate for the ground true expected conditional, where $p\left(\widetilde{Y} \mid X, \mathcal{D}^\text{BM}\right)$ is the posterior predictive and $\mathcal{D}^\text{BM}$ denotes the dataset used for inferring the posterior of the \gls{bm}.

As we are working with a \gls{bm}, the posterior predictive distribution requires marginalizing over the posterior 
\begin{equation}
	p\left(\widetilde{Y} \mid X, \mathcal{D}^\text{BM} \right) = \int p\left(\widetilde{Y} \mid X, \phi \right) p\left(\phi \mid \mathcal{D}^\text{BM} \right) d\phi .
 \end{equation}
As a result, doing risk minimization with a \gls{bm} requires computing the double integral
\begin{equation}
\label{eq:expectation_given_x_phi_paper}
\begin{split}
	\mathbb{E}\left[\widetilde{Y} \mid X \right] &= \int \widetilde{Y} \int p\left(\widetilde{Y} \mid X, \phi \right) p\left(\phi \mid \mathcal{D}^\text{BM}\right) \, d\phi \, d\widetilde{Y} \\
	&= 	\int \mathbb{E}\left[ \widetilde{Y} \mid X, \phi \right] p\left(\phi \mid \mathcal{D}^\text{BM}\right) \, d\phi .
\end{split}
\end{equation}
Integrating out the posterior is in most cases computationally intractable. As an approximation, one could ignore the full distribution over the posterior, and simply use a point estimate of the posterior, this point being the one with the highest probability, $\phi_\text{MAP}$ \cite{goodfellow2016deep}. This would constitute a \gls{map} estimate. With uninformative priors, this would even be the \gls{mle} \cite{mackay2003information}. This estimate will not reflect the model's uncertainty and is \emph{not} a prediction that minimizes risk. Further, using $\phi_\text{MAP}$ instead of marginalizing over the posterior can lead to more extreme predictions \cite{mackay2003information, mcelreath2020statistical}.

Luckily, one can do better than \gls{mle} and \gls{map}. One can obtain an estimate of~\eqref{eq:expectation_given_x_phi_paper} through the simple \gls{mc} estimator \cite{Kroese2014WhyTM, kruger2016predictive}:
\begin{equation}
	\label{eq:mc_sim_paper}
	\begin{split}
		\boldsymbol{\phi} &= \left[ \phi_m \sim p\left( \phi \, \mid \, \mathcal{D}^\text{BM} \right) \right]_{m = 1, \ldots, M}^T \\
		\widetilde{y} &\approx \frac{1}{M} \sum_{m = 1}^{M} \mathbb{E}\left[ \widetilde{Y} \mid X, \phi_m \right] ,\\
	\end{split}
\end{equation}
where $M$ is the number of \gls{mc} samples. Equation~\eqref{eq:mc_sim_paper} can be calculated in parallel across CPU cores as each simulation runs independently of the other. Despite \gls{mc} simulation being a simple, efficient, and fairly accurate, approximation, the degree of parallelism is limited to the number of CPU cores available, and consequently insufficient in time-sensitive domains with the need for a high level of accuracy, i.e.\@ more posterior samples used for prediction. For the most simplistic \glspl{bm}, this is not a computational burden and involves only a few matrix operations. However, as the complexity of the model increases, so does the cost of making predictions. This is particularly the case for \glspl{bm} requiring some form of recursion. 

\begin{figure*}[tb]
	\centering
	\def\postDistColor{red}
\def\dataDistColor{blue}
\def\likeDistColor{darkgray}
\def\priorDistColor{olive}

\def\axisHeight{5cm}
\def\axisToTitleSpace{5mm}
\def\axisWidth{5cm}
\def\spaceBetweenSections{1.5cm}
\def\arrowSpacing{3mm}
\def\ymax{0.7}

\begin{tikzpicture}[font=\normalsize]
%	\draw[
%	help lines,
%	line width=0.1pt,
%	step=0.1,
%	]  (0,-3) grid (19,6);
	
	\coordinate (data_dist_start) at (0,0);
	\coordinate (coord_bm_start) at (\axisWidth+\spaceBetweenSections, 0);
	\coordinate (coord_bm_end) at (\axisWidth+\axisWidth+\spaceBetweenSections, 0);
	\coordinate (coord_nn_start) at (\axisWidth+\axisWidth+\spaceBetweenSections+\spaceBetweenSections, \axisHeight/2);

	\coordinate (obs_end) at (\axisWidth, \axisHeight/2);
	\coordinate (bm_start) at (\axisWidth + \spaceBetweenSections, \axisHeight/2);
	\coordinate (bm_end) at (\axisWidth + \axisWidth + \spaceBetweenSections, \axisHeight/2);
	\path (bm_start) +(-\arrowSpacing, 0) coordinate (to_bm_arrow_end);
	\path (coord_nn_start) +(-\arrowSpacing, 0) coordinate (to_nn_arrow_end);

	\node[circle, fill=blue, anchor=west] at (coord_nn_start) (nn_data_in) {};
	\draw[->, very thick] (obs_end) -- (to_bm_arrow_end);
	\draw[->, very thick] (bm_end) -- (to_nn_arrow_end);

	%%% DATA DIST START %%%%
	
	\begin{axis}[width=\axisWidth,
		height=\axisHeight,
		scale only axis,
		at={(data_dist_start)},
		every axis plot post/.append style={
			mark=none,samples=500,smooth},
		domain=-2:2, ymax=\ymax,
		axis lines=center, % adds arrowheads
		axis x line*=bottom, % no box around the plot, only x and y axis
		axis y line*=left, % the * suppresses the arrow tips
		enlargelimits=upper, % extend the axes a bit to the right and top
		xtick=\empty, % remove xticks and labels
		ytick=\empty, % remove yticks and labels
		xlabel=$\mathcal{D}$,
		every axis x label/.style={
			at={(ticklabel* cs:0.9,9.0)},
			anchor=west,
		},
		]
		\coordinate (data_sample_coor) at (axis cs:0.3, 0);	  
		\addplot[color=blue] {gauss(0,0.55)};
		\node[font=\color{\dataDistColor}] at (axis cs:-1,0.45) {$p(\mathcal{D})$};

	\end{axis}
	
	\node[inner sep=0,outer sep=0, anchor=center] at (\axisWidth/2, \axisHeight + \axisToTitleSpace + 0.2mm) (obs_title) {\large Observations};
	\node[circle, fill=blue] at (data_sample_coor) (d_sample) {};
	%%% DATA DIST END %%%%
	
	%%% POST DIST START %%%%	
	\begin{axis}[width=\axisWidth,
		height=\axisHeight,
		scale only axis,
		at={(coord_bm_start)},
		every axis plot post/.append style={mark=none,domain=-2:3,samples=500,smooth},
		ymax=\ymax,
		axis lines=center, % adds arrowheads
		axis x line*=bottom, % no box around the plot, only x and y axis
		axis y line*=left, % the * suppresses the arrow tips
		enlargelimits=upper, % extend the axes a bit to the right and top
		xtick=\empty, % remove xticks and labels
		ytick=\empty, % remove yticks and labels
		xlabel=$\phi$,
		every axis x label/.style={
			at={(ticklabel* cs:0.9,9.0)},
			anchor=west,
		},
		]
		\addplot[color=\priorDistColor, dashed] {gauss(-1.,1)};
		\addplot[color=\likeDistColor, dashed] {gauss(1,0.55)};
		\addplot[color=\postDistColor] {gauss(0,0.7)};
		
		\node[font=\color{\priorDistColor}] at (axis cs:-1,0.45) {$p(\phi)$};
		\node[font=\color{\likeDistColor}] at (axis cs:2.4,0.55) {$p(\mathcal{D} \mid \phi)$};
		\node[font=\color{\postDistColor}] at (axis cs:0,0.62) {$p(\phi \mid \mathcal{D})$};
		
		\coordinate (posterior_sample_1) at (axis cs:0.15, 0);
		\coordinate (posterior_sample_2) at (axis cs:-0.3, 0);
		\coordinate (posterior_sample_3) at (axis cs:-1., 0);
		\coordinate (posterior_sample_4) at (axis cs:1.5, 0);
		\coordinate (posterior_sample_5) at (axis cs:0.7, 0);
		
	\end{axis}
	
	\draw[inner sep=0,outer sep=0, anchor=center] (\axisWidth+\spaceBetweenSections+\axisWidth/2, \axisHeight + \axisToTitleSpace) + (0, -0.2mm) node (bm_title) {\large Bayesian Model $f$};
	
	\node[circle, inner sep=0,outer sep=0, fill=lightgray] at (posterior_sample_1) (d_sample) {$\textcolor{\postDistColor}{\phi_1}$};
	\node[circle, inner sep=0,outer sep=0, fill=lightgray] at (posterior_sample_2) (d_sample) {$\textcolor{\postDistColor}{\phi_2}$};
	\node[circle, inner sep=0,outer sep=0, fill=lightgray] at (posterior_sample_3) (d_sample) {$\textcolor{\postDistColor}{\phi_3}$};
	\node[circle, inner sep=0,outer sep=0, fill=lightgray] at (posterior_sample_4) (d_sample) {$\textcolor{\postDistColor}{\phi_4}$};
	\node[circle, inner sep=0,outer sep=0, fill=lightgray] at (posterior_sample_5) (d_sample) {$\textcolor{\postDistColor}{\phi_5}$};

	%%% POST DIST END %%%%
	
	%%% NN START %%%
	
	\node[right=1.5cm of nn_data_in, anchor=north, draw, ellipse, rotate=-90] (hidden_layers) {\normalsize Hidden Layers};
	\draw (nn_data_in) -- (hidden_layers);

	\draw (hidden_layers) + (1.5cm, 0) node[fill=\postDistColor, circle] (output_3) {};	
	\node[above=2mm of output_3, fill=\postDistColor, circle] (output_2) {};
	\node[above=2mm of output_2, fill=\postDistColor, circle] (output_1) {};
	\node[below=2mm of output_3, fill=\postDistColor, circle] (output_4) {};
	\node[below=2mm of output_4, fill=\postDistColor, circle] (output_5) {};
	
	\node[right=1mm of output_1, inner sep=0, outer sep=0] (output_1_text) {$\mathbb{E}\left[Y \mid \textcolor{\dataDistColor}{X}, \textcolor{\postDistColor}{\phi_1} \right]$};
	\node[right=1mm of output_2, inner sep=0, outer sep=0] (output_2_text) {$\mathbb{E}\left[Y \mid \textcolor{\dataDistColor}{X}, \textcolor{\postDistColor}{\phi_2} \right]$};
	\node[right=1mm of output_3, inner sep=0, outer sep=0] (output_3_text) {$\mathbb{E}\left[Y \mid \textcolor{\dataDistColor}{X}, \textcolor{\postDistColor}{\phi_3} \right]$};
	\node[right=1mm of output_4, inner sep=0, outer sep=0] (output_4_text) {$\mathbb{E}\left[Y \mid \textcolor{\dataDistColor}{X}, \textcolor{\postDistColor}{\phi_4} \right]$};
	\node[right=1mm of output_5, inner sep=0, outer sep=0] (output_5_text) {$\mathbb{E}\left[Y \mid \textcolor{\dataDistColor}{X}, \textcolor{\postDistColor}{\phi_5} \right]$};
	
	\draw (hidden_layers) -- (output_1);
	\draw (hidden_layers) -- (output_2);
	\draw (hidden_layers) -- (output_3);
	\draw (hidden_layers) -- (output_4);
	\draw (hidden_layers) -- (output_5);
	
	%%% NN END %%%
	
	\draw[inner sep=0,outer sep=0, anchor=center ] (output_1 |- obs_title) + (0,  -0.2mm) node (nn_title) {\large Neural Network $g$};
	%\node[right= of bm_title, fontscale=1] (nn_title) {\large Neural Network $g$};	
	%%%% LIST W PROS AND CONS %%%
	%	\node[below left=\axisHeight+10mm and 0cm of bm_title, anchor=south east]	
	\node[inner sep=0,outer sep=0, below=\axisHeight+8mm of bm_title, fontscale=1] (bm_pros_n_cons) {\normalsize \begin{tabular}{@{}l@{}}$+$ Easy to interpret \\ $+$ Data efficient \\ $+$ Integrated uncertainty \\ $-$ Slow approximations \end{tabular}};
	
	\draw[inner sep=0,outer sep=0, fontscale=1] (nn_title |- bm_pros_n_cons) node (nn_pros_n_cons) {\normalsize \begin{tabular}{@{}l@{}} $-$  Hard to interpret \\ $-$ Data heavy \\ $+$ Flexible \\ $+$ Fast predictions \end{tabular}};

\end{tikzpicture}
	\caption{\textbf{The transition from data, to \gls{bm}, to \gls{nn}.} The colored dots indicate a sample from the corresponding distribution. First, the data is sampled. Then, the \gls{bm} is fitted, and $M$ samples from the posterior distribution is taken. Finally, the \gls{nn} is trained such that for a given observation, the output of the \gls{nn} corresponds to a prediction by the \gls{bm} using each of the $M$ posterior samples.}
	\label{fig:bm_to_nn}
\end{figure*}
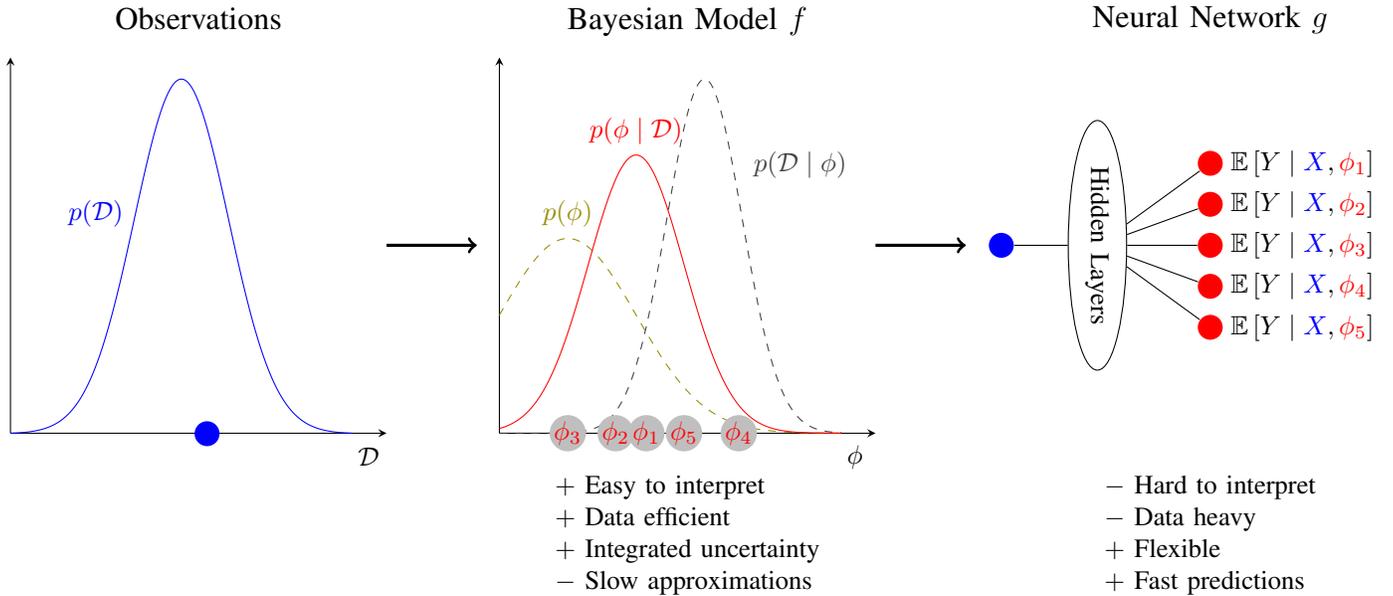

In section~\ref{sec:method}, we present a method for training a \gls{nn} to predict the expectation $\mathbb{E}\left[ \widetilde{Y} \mid X, \phi \right]$ for a set of pre-generated posterior samples. This allows for computing the expectation using the \gls{nn} as a surrogate, and provides a point-wise for $\phi$ approximation of the mean posterior predictive distribution in a single feed-forward pass. The mean of this set of conditional expectations constitutes an \gls{mc} estimate of~\eqref{eq:expectation_given_x_phi_paper}, and can be used for minimizing risk when predicting. The general concept is visualized in Figure~\ref{fig:bm_to_nn}, which shows the transition from observational data, to a fitted BM with distributions over some parameters $\phi$, to a prediction using a \gls{nn} as as surrogate for the BM.

\section{Related Work}
There are several (dis)advantages of using \glspl{bm} over \glspl{nn}.
Generally, \Glspl{bm} allow one to \emph{1)} specify prior beliefs on the parameters to be inferred, \emph{2)} explicitly model interactions between features, and  \emph{3)} make decisions on how these features should affect the predictive distribution \cite{kruschke2014doing, gelman2013bayesian}. This requires a thorough understanding of the data and domain at hand, which can be difficult, if not impossible, in some cases. 
However, this construct results in data-efficient models, capable of fitting even on very sparse datasets \cite{CORREA20097270, goodfellow2016deep}. The use of prior, likelihood, and posterior distribution, results in a strong, consistent, way of handling uncertainty throughout all aspects of the model. This is a result of the extensive use of marginalization over these distributions. This marginalization is what distinguishes the method from other pure optimization methods like \glspl{nn} \cite{Wilson2020}. It is also what makes many interesting \glspl{bm} computationally heavy, as doing exact inference for those is not possible due to the lack of a closed-form solution. Therefore, a variety of approximation tools have been explored to overcome the computational demand for the integrals involved in the marginalization procedure \cite{mcelreath2020statistical, kruschke2014doing, bishop2006pattern}.

The theoretical framework is inherently founded on probability theory. Therefore, the \gls{ml} community has adopted and applied many concepts from Bayesian theory in their work to improve \glspl{nn} \cite{Wilson2020, Maddox2019, Wilson2020_a, tran2017bayesian, pearce2020uncertainty, myshkov2016posterior, bishop1995bayesian, neal2012bayesian, hernandezlobato2015probabilistic, NIPS2011_7eb3c8be}. Bishop \cite{bishop1995bayesian} performed a thorough walk-through of some of these adopted concepts, including model comparison, hyperparameter optimization, \gls{al}, ensemble methods, and $L_2$ regularization \cite{mackay1992bayesian, bishop1995bayesian, bishop2006pattern}.

It would seem that \glspl{nn} have little to contribute to Bayesian methods. However, one of the main assets of \glspl{nn} is their extreme flexibility as they are universal approximators \cite{hornik1991approximation, sonoda2017neural, goodfellow2016deep}. This is a benefit of the vast amount of free parameters in the \glspl{nn}. The number of free parameters of a \gls{nn} is often several orders of magnitude higher than most \glspl{bm}. Taking for instance GPT-3 which uses 175 billion parameters \cite{brown2020language}. The large number of free-parameters in \glspl{nn} gives them a high model capacity. A model with a high capacity is flexible, but dependent on large quantities of training data to prevent overfitting. The \gls{vc} dimension can be used to obtain an upper bound on the generalization loss of a model given its effective capacity and the size of the training dataset \cite{vapnik1999overview}. As the capacity of the model increases, the required size of the training dataset increases to maintain the same generalization performance \cite{bishop2006pattern, goodfellow2016deep, sontag1998vc}. \glspl{nn} with millions of free parameters are, as a result, depending on large training datasets. Finally, \glspl{nn} generally suffer a lack of interpretability compared to a \gls{bm}. The inferred \gls{mle} of the free parameters of a \gls{nn} are simply far more difficult to reason about and interpret compared to a \gls{bm}. Despite having far more free parameters, their extended use of matrix multiplication, execution on GPUs, and lack of marginalization, make them very fast compared to \glspl{bm}.

Recent development in the field of \gls{ml} applies the Bayesian probabilistic framework and one or more \glspl{nn} in a compound for doing variational inference \cite{blei2017variational} of complex distributions. Examples of such approaches are generative models like \acrlongpl{vae} \cite{Kingma2014} and normalizing flows \cite{papamakarios2021normalizing, rezende2015variational}. For instance, normalizing flows can be used for approximating the posterior distribution of a Bayesian \gls{nn}\cite{louizos2017multiplicative}.

For such generative models, the posterior distribution and \gls{nn} are optimized jointly and thus tightly coupled. The composition leverage the flexibility of \glspl{nn} and the principled probabilistic framework to infer distributions over data.

Another type of combining Bayesian methods and \glspl{nn} is fitting a \gls{nn} to the posterior distribution of a \gls{bm}. In such setups the posterior distribution of the \gls{bm} is first inferred, and only afterwards the \gls{nn} is introduced. This is done with the purpose of leveraging the predictive speed of the \gls{nn}, and such an approach is also presented in this paper.

A pre-existing example of this method is presented by Jia et al. \cite{jia2017using}. In which case, the \gls{nn} predicts a discrete approximation to the conditional posterior distribution for each free parameter in the \gls{bm}. This is fundamentally different from our approach, as our \gls{nn} has no notion of the parameters of the \gls{bm}, it is simply latent in the dataset.

In the work of Pavone et al.\@~\cite{Pavone_2019}, another approach was taken to approximate a \gls{nn} to a \gls{bm}. In this work, they took the domain of particle physics and sampled a target $t \sim p(T)$ using a joint prior distribution over the target variables. Then, using the likelihood, $I$ samples of datapoints $\mathbf{x} = \left[x_i \sim p(X \mid t) \right]^T_{i = 1,\ldots,I}$ served as $I$ input examples to the \gls{nn}. The \gls{nn} would then gain robustness and learn the uncertainty of the model through the data sampling process. The \gls{nn} itself though, predicts a single \gls{mle} for $t$, and thus has no way to represent the posterior predictive distribution in its output. The results show how using a \gls{nn} for predictions rather than a \gls{bm}, can reduce the prediction time from over four hours to a matter of milliseconds. This demonstrates the potential benefits from using such setups \cite{Pavone_2019}.

\section{Method}\label{sec:method}
In this section, we introduce the method and algorithms proposed for approximating a \gls{bm} with a \gls{nn}. The purpose of the approximation is to have the ability to make predictions on new data based on an approximation of the full posterior distribution of the \gls{bm}. These predictions should \emph{minimize the risk} one takes when making that prediction -- meaning it should be the best possible prediction one can make given a \gls{bm}. In the result section, we show how our method allows for such predictions faster than standard methods for \glspl{bm} with higher complexity. This happens due to the architecture as presented in Figure~\ref{fig:bm_to_nn}. The output of the \gls{nn} is a \emph{point-wise for $\phi$ estimate of the \gls{bm}'s mean posterior predictive distribution}. This approximation is obtained in a \emph{single} feed-forward pass of the \gls{nn}. Having access to the posterior predictive is beneficial for subsequent evaluation of the model as it provides more detailed information about the predictions. This distribution can, for instance, be used for quantifying the model's uncertainty in the prediction made. 

One of the major disadvantages of \glspl{nn} is their requirement for large training datasets in contrast to \glspl{bm}. This issue is resolved with the ability to use the \gls{bm} for generating a dataset of arbitrary size for the \gls{nn}. Having a set of posterior samples from the \gls{bm}, one can simply generate synthetic input data, and pass it through the \gls{bm}. This data needs to be evaluated on the \gls{bm} multiple times, one time for each posterior sample used. The result is a discrete approximation of the mean of posterior predictive distribution. 
It is important to note that the \gls{nn} has no notion of the actual posterior samples of the \gls{bm}. These are latent in the dataset generated for training the \gls{nn}.
In this way, our method preserves the benefits from the \gls{bm}, but also gain the speed improvement from the \gls{nn}.

Formally, we let $g : \mathbf{x} \subset \mathbb{R}^J \rightarrow \widetilde{\mathbf{y}} \subset \mathbb{R}^M$ be our approximation of the model (i.e.\@ our \gls{nn}). For fitting $g$, we apply the loss function $\text{Smooth}_{\text{L1}}(\mathbf{y}, \widetilde{\mathbf{y}})$, with $\mathbf{y} = \left[ y_m = \mathbb{E}\left[Y \mid \mathbf{x}, \phi_m \right]\right]_{m = 1, \ldots, M}^T$. $\boldsymbol{\phi}$ is a vector of size $M$ with $\phi \sim p(\phi \mid \mathcal{D}^\text{\gls{bm}})$ sampled prior to data generation and afterwards kept fixed\footnote{As the conditional expectation of the \gls{bm} is now the ground true value to be predicted by the \gls{nn}, the output of the \gls{bm} is now denoted $\mathbf{y}$ and the output of the \gls{nn} as $\mathbf{\widetilde{y}}$.}. For computing the expectations, the posterior predictive, $p\left( Y \mid X, \phi\right)$, from the \gls{bm} is used. Finally, $\text{Smooth}_{\text{L1}}$ is the Smooth L1-loss, a combination of L1-loss and L2-loss \cite{ren2015faster}.

The core algorithm contains three steps: \emph{1)} Let $\boldsymbol{\phi}$ be a vector of $M$ posterior samples taken from $p\left(\phi \mid \mathcal{D}^\text{\gls{bm}} \right)$, \emph{2)} sample the training dataset, $\mathcal{D}^\text{NN}$, using algorithm \ref{algo:DataSampling}, and \emph{3)} fit $g$ to $\mathcal{D}^\text{NN}$.

\begin{algorithm}[tb]
\DontPrintSemicolon
\LinesNumbered
\SetAlgoLined
\Input{$I$, $p\left(X\right)$, $\tau$, $p\left(Y \mid X, \phi \right)$}
\Output{$\mathcal{D}^\text{NN}$}   
$\mathcal{D}^{\text{NN}} \gets \varnothing$ \;
\For{$i\gets0$ \KwTo $I$ \KwBy $1$}{
	$\boldsymbol{\rho} \gets \left[ \, \rho_{j} \sim \text{Bern}\left(\tau\right) \right]_{j = 1, \ldots, J}^T$ \;
	$\mathbf{x} \sim p\left(X \right)$\;
	$\mathbf{x} \gets \mathbf{x} \odot \boldsymbol{\rho}$ \hfill \tcp{Dropout}
	$\mathbf{y} \gets \left[ y_m \gets \mathbb{E}\left[Y \mid \mathbf{x}, \phi_m \right]\right]_{m = 1, \ldots, M}^T$\;
	$\mathcal{D}^{\text{NN}} \gets \mathcal{D}^{\text{NN}} \cup \{\left( \mathbf{x}, \mathbf{y}\right) \}$\;	
}
 \caption{Algorithm for the data sampling process for fitting the \gls{nn} to a \gls{bm} for risk minimization.}
 \label{algo:DataSampling}
\end{algorithm}

The sampling algorithm is outlined in Algorithm~\ref{algo:DataSampling}. Here, $\odot$ denotes the Hadamard product. We randomly \emph{dropout} with probability, $\tau$, each of the $J$ sampled values in vector $\mathbf{x}$. This is done as a generalization technique to assist the \gls{nn} recognizing invariants in the underlying model. This is similar to randomly removing pixels for an image recognition task, only in our case, the predicted value, $\mathbf{y}$, is changed with respect to the alteration of $\mathbf{x}$.

\subsection{Computational Complexity}
\label{sec:computational_complexity}
Generating data and fitting a \gls{nn} to a \gls{bm} is a computationally more time-consuming task compared to doing a \gls{mc} simulation for a single prediction. If each \gls{bm} prediction with MC has complexity $\mathcal{O}(m)$, with $m$ being the number of samples, then making $n$ predictions has complexity $\mathcal{O}\left(nm\right)$. As the number of predictions increases, the processing time increases linearly. In contrast, generating a dataset $\mathcal{D}^\text{NN}$ and doing $n$ predictions has complexity $\mathcal{O}\left(\kappa m + n\right)$ with $\kappa$ being the number of dataset samples required for training the \gls{nn}. When $n \geq \frac{\kappa m}{m -1}$, our method has the lowest overall complexity. One might wonder if that is a real use-case, as $\kappa$, is usually in the order of thousands. Some optimization problems and domains, however, are depending on solving the conditional expectation, $\mathbb{E}\left[Y \mid X \right] $, a vast amount of times for varying input. Moreover, for a real-time application, once trained, the \gls{nn} will provide better response times and thus improve user experience.

Having a fitted \gls{nn} and $m > 1$, using the \gls{nn} allows for risk minimized predictions with a lower complexity for a single observation compared to the \gls{bm}. This makes our method appealing in some time-sensitive domains. These are domains with surplus time when fitting the \gls{bm}, but at some point later require accurate predictions quickly for new input. Examples of such domains could be stock market trading or autonomously driving vehicles: both needing rapid, correct, reactions to new observations, but have surplus time when the stock market is closed, or doing research. Further, our method is beneficial when the rate of new incoming data surpasses the time it takes to evaluate~\eqref{eq:mc_sim_paper} using the \gls{bm}. Examples are particle physics and big data applications with a high data velocity.

Some software frameworks such as Jax \cite{jax2018github}, enable executing \glspl{bm} on the GPU rather than the CPU. This provides some gain in speed. However, the computational complexity analysis remains the same as the same sequence of statements needs to be computed for the \gls{bm}.

\subsection{Active Learning}
\label{sec:al}
As the computational complexity of using our method depends on $\kappa$ (the size of the training set), minimizing $\kappa$ increases the computational benefit of our proposed method. Thus, we apply \gls{al} as an extension to the regular training algorithm to maximize the computational benefit. The approach is outlined in algorithm \ref{algo:AL}, and works by iteratively expanding the training dataset for the \gls{nn}. 
\begin{algorithm}[tb]
	\DontPrintSemicolon
	\LinesNumbered
	\SetAlgoLined
	\Input{$\boldsymbol{\phi}$, $\tau$, $p\left(Y \mid X, \phi \right)$, $p\left(\phi \mid \mathcal{D}^\text{BM}\right)$, patience}
	\Output{$g$}
	stopper $\gets$ EarlyStopping(patience) \;
	$X \sim \mathcal{U}\left(\mathbf{0}_J, \mathbf{1}_J\right)$ \;
	$g \gets$ initialize NN \;
	$\boldsymbol{\phi} \gets \left[ \phi_m \sim p\left( \phi \, \mid \, \mathcal{D}^\text{BM} \right) \right]_{ m = 1, \ldots, M}^T$\;
	$\mathcal{D}^{\text{NN}} \gets$ algorithm \ref{algo:DataSampling}($I^\text{Init}$, $\boldsymbol{\phi}$, $p\left(X\right)$, $\dots$)\;
	\Do{$\mathrm{stopper.should\: continue}\left(g\right)$}{%
		$\mathcal{D}^{\text{NN}} \gets \mathcal{D^{\text{NN}}} \: \cup $ algorithm \ref{algo:DataSampling}($I^\text{AL}$, $\boldsymbol{\phi}$, $p\left(X\right)$, $\dots$) \;
		$g \gets$ (re)train $g$ on $\mathcal{D}^{\text{NN}}$ \;
		$\boldsymbol{\sigma} \gets \text{MeasureUncertainty}\left(g\right)$ \;
		$\boldsymbol{\pi} \gets \text{Softmax}\left(\boldsymbol{\sigma} \right)$\;
		$X \sim \text{Cat}\left(\boldsymbol{\pi} \right)$ \;
	}
	\caption{Active Learning algorithm.}
	\label{algo:AL}
\end{algorithm}
This is done through modifications to the data sampling distribution, $p\left(X\right)$, from which new data is generated. The iterative, data-augmented, approach helps minimizing the computational burden of evaluating the \gls{bm}. During the first iteration, a uniform distribution is used for generating the initial dataset of size $I^\text{Init}$. Subsequently, this distribution is changed to a categorical distribution with probabilities reflecting the \gls{nn}'s predictive \emph{uncertainty} on entries from a large, unlabeled, uniformly sampled, dataset $\mathbf{X}^\text{uncert.}$. From this categorical distribution, $I^\text{AL}$ entries are sampled with replacement. These are then labeled using the \gls{bm} and added to the training dataset for the next round.

The aforementioned uncertainty is measured through the use of dropout. For each example $\mathbf{x}^\text{uncert.} \in \mathbf{X}^\text{uncert.}$, we measure the \gls{nn}'s predictive uncertainty. This is done by performing $K$ feed-forward passes through the network for $\mathbf{x}^\text{uncert.}$ with dropout enabled, such that the predictions, $\widetilde{\mathbf{Y}}^\text{uncert.}$, is a $M \times K$ matrix. To reduce this matrix to a single value reflecting uncertainty on $\mathbf{x}^\text{uncert.}$, we take the standard deviation over $K$ to obtain a vector of $M$ standard deviations for which we calculate the mean: 
\begin{equation}
\label{eq:nn_uncert}
	\sigma_{\mathbf{x}^\text{uncert.}} = \frac{1}{M} \sum_{m \in M} \sigma \left( \widetilde{\mathbf{Y}}^\text{uncert.}_{m} \right)
\end{equation}
The larger standard deviation means a higher uncertainty and as a result a larger sampling probability from the categorical distribution.
This operation is performed over each instance in the unlabeled dataset, and the resulting set of $\boldsymbol{\sigma}$'s is transformed using Softmax in order for them to be used as probabilities.

Finally, we apply \gls{es}~\cite{Prechelt1998} at two levels. We will refer to these as \emph{intra}- and \emph{inter}-\gls{es}. Intra-\gls{es} regulate when to stop fitting the \gls{nn} on $\mathcal{D}^{\text{NN}}$, while inter-\gls{es} determines when to stop the \gls{al} iterations. When inter-\gls{es} chooses to stop, it restores the state of the \gls{nn} to the best performing one. Both \glspl{es} are using the same, fixed, validation dataset to ascertain the performance of the \gls{nn}.

\section{Experiments}
For the experiments in this paper, we assume a Bayesian regression model, $f : \mathbf{x} \subset \mathbb{R}^{J} \rightarrow y \subset \mathbb{R}$, where $\mathbf{x}$ is a single example. The model is as follows
\begin{equation}
\label{eq:var_complex_model}
\begin{split}
	\boldsymbol \alpha \sim \mathcal{N}\left(1.5,\, \mathbf{I}_J\right); &\qquad
	\boldsymbol \beta \sim \mathcal{N}\left(0.5,\, 0.25 \mathbf{I}_J\right) \\
	\sigma^2 \sim \mathcal{N}\left(0,\, 1\right); &\qquad
	\gamma \sim \mathcal{N}\left(0,\, 0.5\right) \\
	f\left(\mathbf{x}\right) = \gamma + \sum_{j = 1}^{J} \beta_j \psi\left(x_j \alpha_j \right); &\qquad
	Y \sim \mathcal{N}\left(f\left(\mathbf{x}\right),\, \sigma^2 \right)	
\end{split}
\end{equation}
With $\mathbf{I}_J$ being the identity matrix with rank $J$.
 The intuition behind this model is that the output is the sum of a linear transformation of each feature which is altered by some function, $\psi$. This function is domain-dependent, and could be $\surd$, $\sin$, $\log$, sigmoid, or the like. $\bm{\alpha}$, $\bm{\beta}$ and $\bm{\sigma}$ are $J$-dimensional column vectors and represent the parameters of the \gls{bm} which needs to be inferred. The \gls{bm} outlined above does not directly resemble any real-world model. Nonetheless, one can easily imagine such a setup for modeling periodic data as the sum of $\sin$ functions with each feature having a different period. Another example could be a model for diagnosing the risk of having a particular disease, here, $\mathbf{x}$ could be a patient's blood test results, $\boldsymbol{\beta}$ the impact of each examined property, $\psi$ the sigmoid function to saturate the influence of each property, and $\boldsymbol{\alpha}$ a scaling factor.
 
Next, we are interested in using the model for predictions over the full posterior on new data. Further, we will assume no closed-form solution of the posterior predictive distribution, hence for doing predictions with the \gls{bm}, we will apply \gls{mc}-sampling as in \eqref{eq:mc_sim_paper}, and assume a sufficient posterior fit to the observed data $\mathcal{D^{\text{BM}}}$.
 
 When conducting predictions using this model, each of the $m \in \{1, \ldots, M\}$ posterior samples of $\bm{\alpha}$, $\bm{\beta}$, and $\gamma$ need to be used.
 The computational complexity of \eqref{eq:var_complex_model} is linearly dependent on $J$. Thus, doing predictions over the full posterior becomes increasingly computationally heavy as the number of features grows.
 One could imagine other terms that reflect more complex effects such as synergies across features to take place in the model. A synergy effect between all features would transform the linearly growing complexity into an exponential one.

In contrast to the complexity growth of the \gls{bm} for increasingly complex tasks, the same growth is not necessarily present for NNs. Using the property of \glspl{nn} being universal approximators, adding more complexity to the \gls{bm} does not necessarily require a larger \gls{nn} as long as its capacity is not fully utilized. Even using a \emph{single} hidden layer in the \gls{nn}, with enough width, it remains a universal approximator \cite{nielsen2015neural}. Consequently, in the extreme case, the \gls{nn} approximation provides extremely fast predictions, as it will only involve two matrix multiplications, two addition terms, and an activation function.
 
 To examine this, our experiments will focus on approximating the \gls{bm} from \eqref{eq:var_complex_model} with a \gls{nn}. We will vary the computational complexity of the model through $J$ and evaluate the \gls{nn}'s fit to the \gls{bm}. The main motivational factor for introducing a \gls{nn} as an approximation of a \gls{bm} is the gain of speed when making multiple predictions. To quantify this, we measure the runtime for making predictions on the whole testing dataset using $M$ posterior samples. We will refer to this runtime as prediction time. Our results will focus on the prediction time as a function of model complexity, and the required size of the training dataset.
 
Throughout all experiments, we use a simple, feed-forward \gls{nn} with a single hidden, dense, layer with a width of $5{,}000$. We use dropout~\cite{srivastava2014dropout} with a rate of $0.5$, and batch normalization~\cite{ba2016layer} after the hidden layer. We let $I^\text{Init} = 10{,}000$, $I^\text{AL} = 1{,}000$, $\tau = 0.8$, and $M = 2{,}000$. We keep $M$ fixed for all experiments and instead vary $J$, but one could make similar experiments for an increasing number of posterior samples. For training the \gls{nn} we use a learning rate of $3 \times 10^{-4}$, and set the patience to $10$ and $20$ for inter- and intra-\gls{es}, respectively. The validation and testing dataset each contain $50{,}000$ examples. The experiments are executed on an Intel Core i7-6700K with 48GB ram and a NVIDIA GeForce RTX 2080TI GPU. We use PyMC3 \cite{salvatier2016probabilistic} version 3.9.3 for Bayesian inference with the No-U-Turn Sampler \cite{hoffman2014no} using $2{,}000$ warmup steps and $2{,}000$ samples. $\mathcal{D}^\text{BM} = \{\mathbf{X}^\text{BM},  \mathbf{Y}^\text{BM}\}$ contains $N = 5{,}000$ examples with $\mathbf{X}^\text{BM}$ being sampled from a standard Gaussian distribution, and $\mathbf{Y}^\text{BM} = \left[\mathbf{Y}^\text{BM}_n = f\left(\mathbf{X}^\text{BM}_{n}\right)\right]^T_{n=1,\ldots,N}$ from \eqref{eq:var_complex_model} with the ground-true values pre-sampled using $\widehat{\gamma} \sim \mathcal{U}\left(-0.5, 0.5\right)$, $\boldsymbol{\widehat{\alpha}} \sim \mathcal{U}\left(0.3 \cdot \mathbf{1}_J, 3.0  \cdot \mathbf{1}_J \right)$, and $\boldsymbol{\widehat{\beta}} \sim \mathcal{U}\left(0.1 \cdot \mathbf{1}_J, \mathbf{1}_J\right)$.
 
\section{Results}
\begin{figure}[tb]
	\centering
	\includegraphics[width=\linewidth]{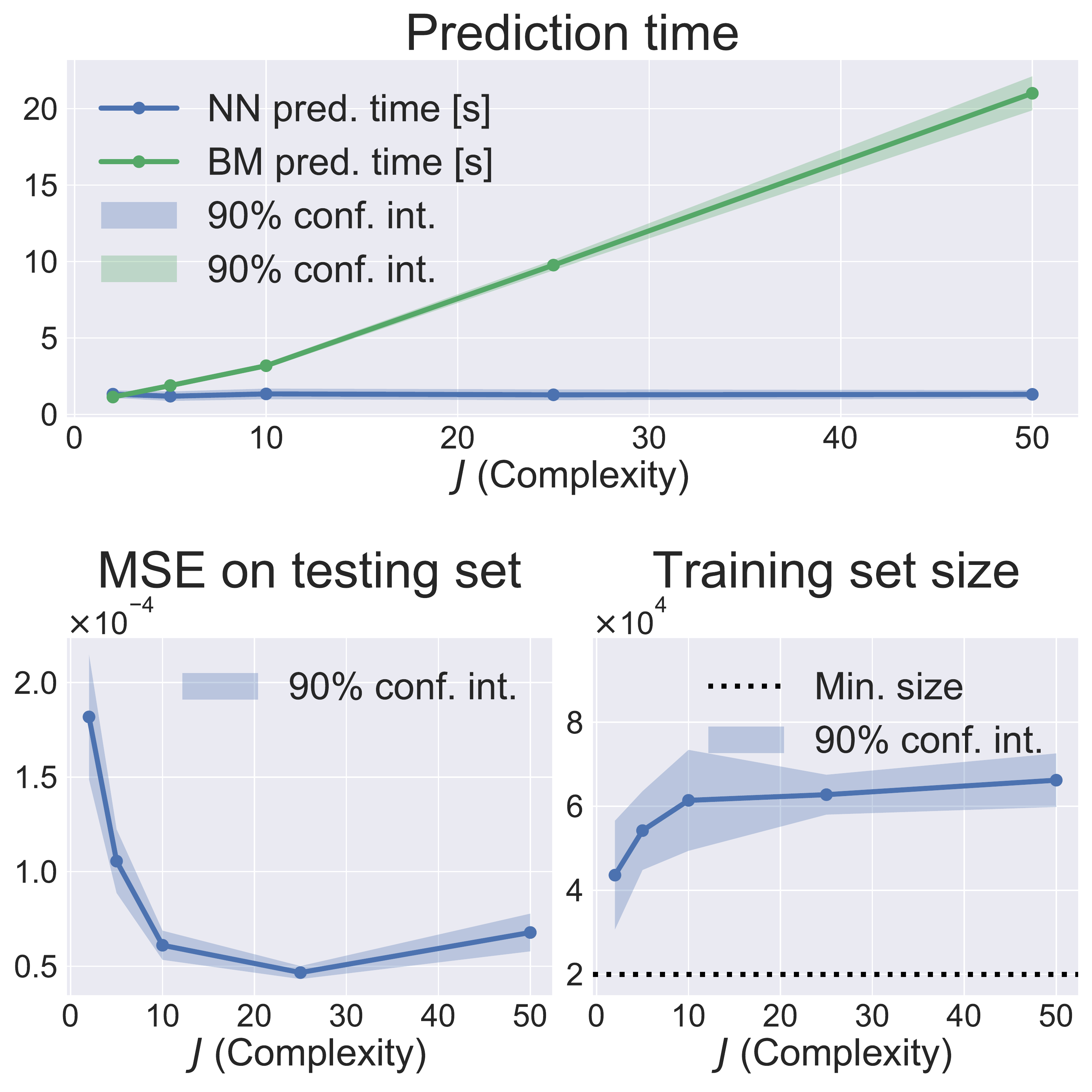}
	\caption{\textbf{Prediction time (top), \gls{mse} on the testing dataset (bottom left), and the ultimate size of the training dataset (bottom right), shown as a function of model complexity $J$.} The lines indicate the mean of each experiment which is repeated five times. The shades are the 90\% confidence interval bands, and the markers show the values of $J$ for which experiments were conducted. The prediction time on the testing dataset using the \gls{bm} versus the \gls{nn} is shown at the top figure. This figure illustrates the linear relation between the model complexity and the prediction time using the \gls{bm} while being constant for the \gls{nn}. The bottom left figure shows the \gls{mse} of the \gls{nn} calculated on the testing dataset. The bottom right figure shows how the size of the training dataset increases nonlinearly with the complexity of the \gls{bm} to be approximated. Here, the dotted line indicates the lowest possible size of the training dataset as each experiment starts with $10{,}000$ examples and passes at least 10 \gls{al} iterations.}
	\label{fig:Speedtest}
\end{figure}

In Figure~\ref{fig:Speedtest}, one can see the prediction time on the testing set for increasing model complexities, the corresponding \gls{mse}, and the size of the training dataset used for fitting the \gls{nn}. When measuring the prediction time of the \glspl{bm}, the operation is run in parallel using all cores of the CPU when iterating over the $M$ posterior samples. This is done using a pool of threads. All calculations make use of the Numpy library with vectorized operations to obtain the best possible performance. The PyMC3 framework also provides a method for making predictions using vectorized computations. However, this does not make use of parallelization and therefore yields a worse prediction time than those we have reported here using both vectorized operations and concurrency.

Measuring runtimes on modern computers is generally an error-prone task, and we, therefore, repeated each experiment five times. From the figure the linear relationship between complexity and prediction time using the \gls{bm} is clear, whilst it remains constant for the \gls{nn}. As complexity increases, we see a small decreasing tendency in the \gls{mse}. The \glspl{mse} is for all experiments considered low, spanning from $4.3 \times 10^{-5}$ to $2.1 \times 10^{-4}$. The results indicate a negative correlation between the \gls{mse} and model complexity. We find this somewhat surprising, as it seems intuitive that more complex models would be harder for the \gls{nn} to approximate. We attribute this phenomenon to be a result of the central limit theorem \cite{polya1920zentralen}, as the normalized distribution over $\mathbf{Y}^\text{NN}$ shrinks to a normal, making it easier for the \gls{nn}.

\subsection{Active Learning and the Size of the Training Dataset}

\begin{figure}[tb]
	\includegraphics[width=\linewidth]{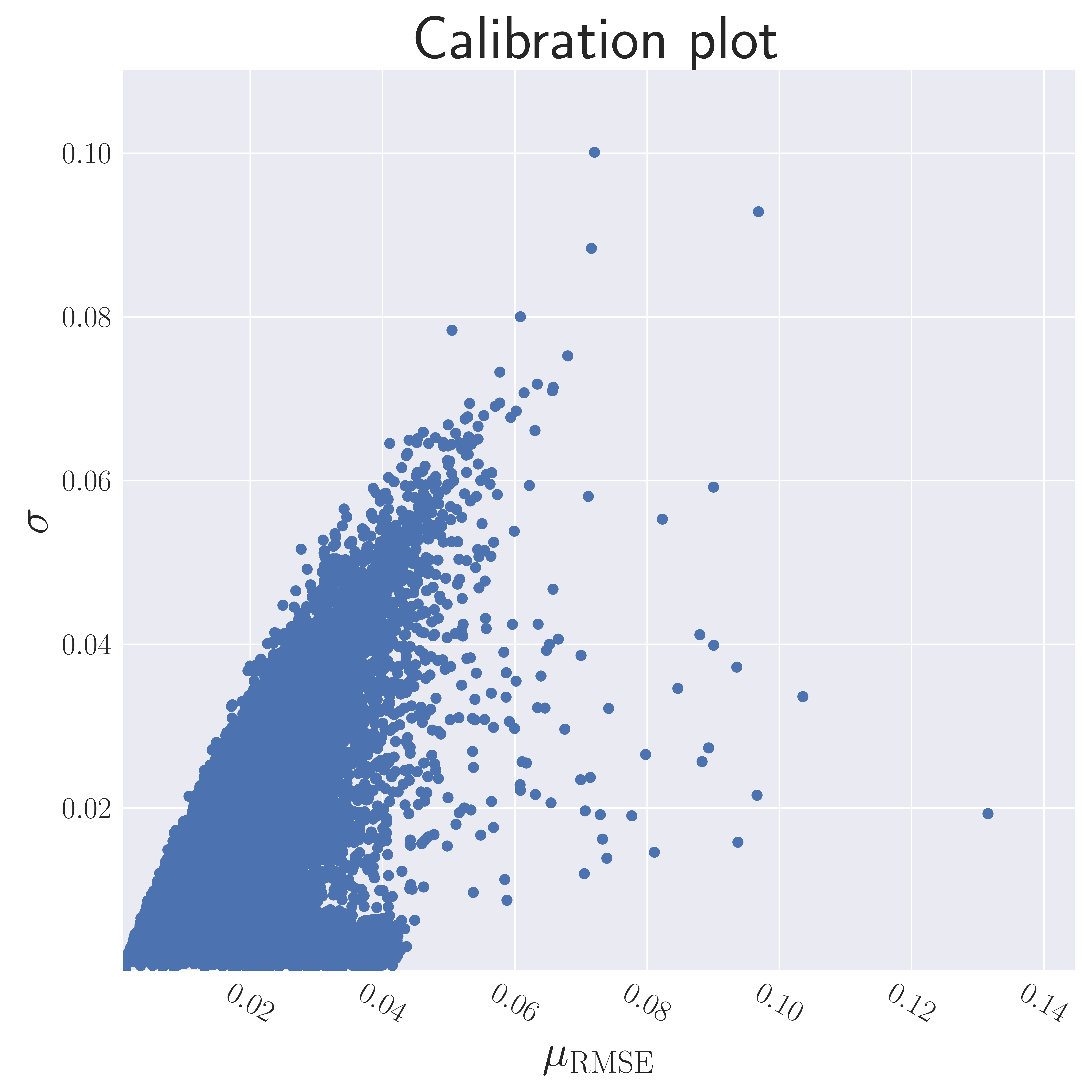}
	\caption{\textbf{Calibration plot showing the correlation between the uncertainty of the \gls{nn}, $\sigma$, and its prediction's \gls{rmse} on newly sampled data, $\mu_\text{RMSE}$.} The correlation allows for sampling additional data purely based on the uncertainty estimate of the datapoints.
	}
	\label{fig:CalibrationPlot}
	\centering
\end{figure}

To examine the strength of the correlation between the \gls{nn}'s uncertainty and the predictive error on the $\mathbf{X}^\text{uncert.}$ dataset, a calibration plot is shown in Figure~\ref{fig:CalibrationPlot}. This figure shows the \gls{nn}'s uncertainty (calculated using~\eqref{eq:nn_uncert}) as a function of the average \gls{rmse}, $\mu_\text{RMSE}$. This average is taken over the $K \times M$ predictions.
The figure shows the correlation for an experiment with complexity $J=5$. Although the correlation is not equally strong for all examples, it shows a general tendency. If the correlation was not present, the newly added data to the training dataset would provide little to no improvement on the validation dataset. If that was the case, the \gls{es} algorithm would terminate the training process as soon as the iteration counter reached the patience threshold. This level is indicated by the dotted line in Figure~\ref{fig:Speedtest}. The line illustrates the smallest possible size of the training dataset ($20{,}000$) when using the hyperparameters $I^\text{Init} = 10{,}000$, $I^\text{AL} = 1{,}000$, and an inter-\gls{es} patience of $10$.

\section{Discussion and Future Work}
The use of dropout as a measure of uncertainty assumes a correlation between the standard deviation over the $K$ predictions, and the actual ground true error.
\cite{Tsymbalov2018} showed such a correlation exists for a range of different problems for which $\widetilde{y} \subset \mathbb{R}^1$. In our domain presented here, we have increased complexity as $\widetilde{\mathbf{y}} \subset \mathbb{R}^M$ with $M \gg 1$. In \eqref{eq:nn_uncert} we take the mean over $M$ to obtain a single scalar reflecting the \gls{nn}'s joint uncertainty over all predicted values. Whether this is sufficient summary statistics for the use of \gls{al} is unsettled, and an area for future research.

In this paper, we presented a method for which the \gls{nn} predicts a vector of point-wise estimations of the \gls{bm}'s posterior predictive distribution $\mathbb{E}\left[\widetilde{Y} \mid X, \boldsymbol{\phi} \right]$.
One could potentially simplify the method by making the \gls{nn} predict $\mathbb{E}\left[\widetilde{Y} \mid X \right]$ instead. Such a change would mean $M=1$, and only require an alteration to the generation of $\mathcal{D}^\text{NN}$ by performing the averaging step in \eqref{eq:mc_sim_paper} when computing $\mathbf{Y}^\text{NN}$ using the \gls{bm}. We argue that one might as well do the MC simulation step on the output of the \gls{nn} rather than when generating the training data for the \gls{nn}. The MC simulation involves a summation operation, and thus information to the \gls{nn} is lost. Therefore, we argue the training task of the \gls{nn} simply becomes easier. Further, having a discrete approximation of the posterior predictive distribution, allows one to recover the uncertainty of the \gls{bm} in a way the summation function would make irrecoverable.

The \gls{bm} presented in \eqref{eq:var_complex_model} is relatively simplistic. More complex effects such as synergies across features or an autoregressive component can significantly increase the prediction time of a \gls{bm}. Future research should extend the \gls{bm} with such effects and evaluate the \gls{nn}'s ability to approximate such constructs, as these leave room for even larger speed-gains over the \gls{bm}.

A drawback of our method is the assumption the \gls{bm} is fixed. Changes to the \gls{bm}'s construct or posterior distribution would require a refit of the \gls{nn}. Depending on the magnitude of the change in the \gls{bm}, one might achieve good results by using transfer learning \cite{torrey2010transfer, goodfellow2016deep}, by transferring the state of the original NN to fit the updated \gls{bm}.

\section{Conclusion}
In this paper, we presented a method for training a \gls{nn} as a function of $\mathbf{x}$ to predict a point-wise approximation of the \gls{bm}'s posterior predictive distribution conditioned on $\mathbf{x}$. This is achieved in a single feed-forward pass of the \gls{nn}. We applied this approach on a set of generalized \glspl{bm}, and showed how the fitted \glspl{nn} achieved good approximations. We further evaluated the runtime when making predictions for each of the \glspl{bm} involved. These predictions were conducted on the testing dataset using $M$ posterior samples. Here, we found a linear relation between the complexity of the \gls{bm} and the runtime, while the runtime for making predictions using the \gls{nn} remained unchanged. Thus, we found our method to make predictions on the testing dataset $>\!14\times$ faster than the largest evaluated \gls{bm} -- this with an imperceptible \gls{mse}. In addition, we show our method to be only slightly slower for the most trivial \glspl{bm}. Hence, the proposed method is applicable when the same, complex, \gls{bm} is needed for making predictions in time-sensitive domains.
We also demonstrated how the use of \gls{al} can help reduce the dataset required to train the \gls{nn} in such a context. This method minimizes the cost of obtaining this training dataset. In this regard, our results indicate that the size of the training dataset increases with a smaller constant factor than the increased complexity of the \gls{bm}. This advocates the use of this method, as doubling the complexity of the \gls{bm} only requires slightly larger training datasets.

\section*{Acknowledgments}
We would like to express our very great appreciation to Mr. Miguel González and Dr. Jes Frellsen for their valuable and constructive suggestions and feedback during the development of this research work.
\clearpage
\bibliography{references}

\end{document}